\documentclass[10pt,twocolumn,letterpaper]{article}

\usepackage{iccv}
\usepackage{times}
\usepackage{epsfig}
\usepackage{graphicx}
\usepackage{amsmath}
\usepackage{amssymb}
\usepackage{float}
\usepackage{algorithm}
\usepackage{algorithmic}
\usepackage{balance}
\usepackage{color}

\usepackage{subfig}

\usepackage[pagebackref=true,breaklinks=true,letterpaper=true,colorlinks,bookmarks=false]{hyperref}

\iccvfinalcopy  

\def\iccvPaperID{2327} 

\ificcvfinal\fi

\begin{document}

\title{Segment-Phrase Table\\for Semantic Segmentation, Visual Entailment and Paraphrasing}

\author{Hamid Izadinia$^{\dag}$  \hspace{1.2cm} Fereshteh Sadeghi$^{\dag}$ \\ \hspace{0.4cm}  Santosh K. Divvala$^{\ddag,\dag}$ \hspace{0.6cm}   Yejin Choi$^{\dag}$ \hspace{1.2cm} Ali Farhadi$^{\ddag,\dag}$\hspace{1.4cm}   \\
\\
\hspace{0.4cm}$^\dag$University of Washington\hspace{2.4cm} $^\ddag$The Allen Institute for AI \\
\tt\small \{izadinia,fsadeghi,yejin\}@cs.uw.edu \hspace{0.8cm} \{santoshd,alif\}@allenai.org\hspace{0.5cm}
}

\twocolumn[{ %
\renewcommand\twocolumn[1][]{#1} %
\maketitle
    \vspace*{-8mm}
\begin{center}
    \centering
\includegraphics[scale=0.31]{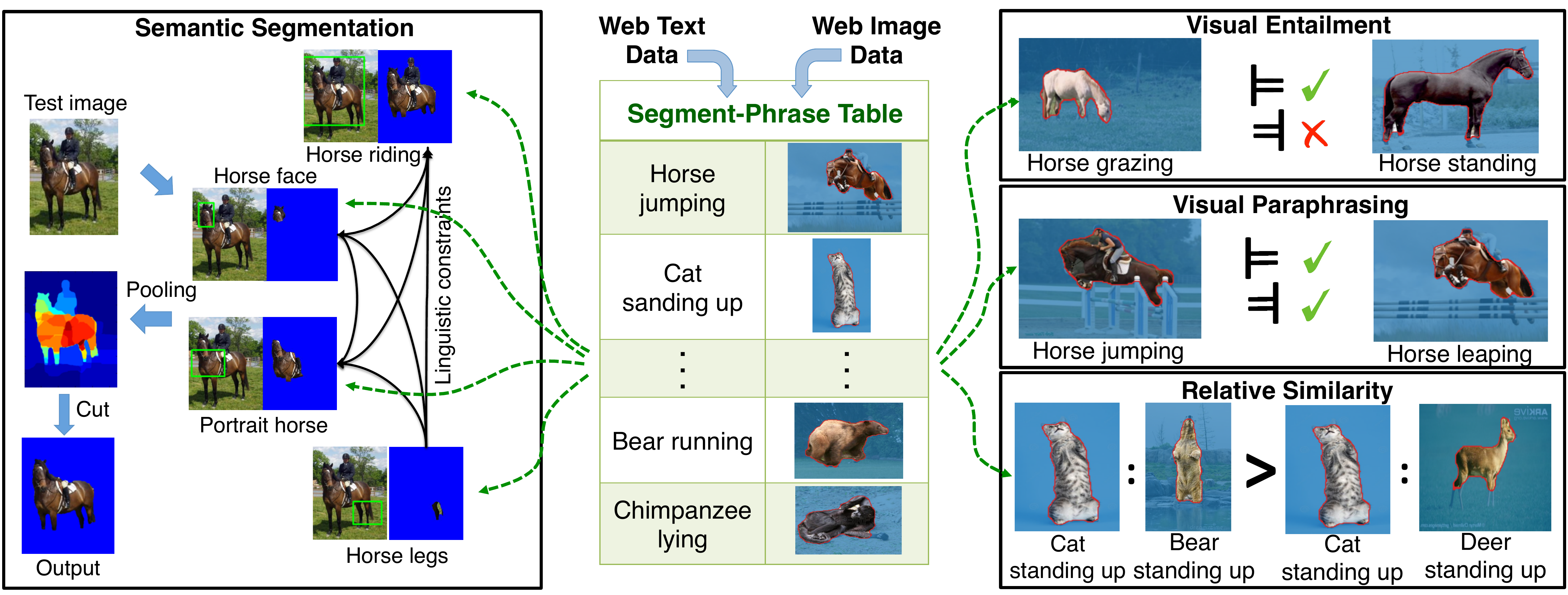}
    \vspace*{-2mm}
    \captionof{figure}{This paper introduces segment-phrase table as a bi-modal resource of correspondences between textual phrases and visual segments. We show that this table facilitates richer semantic understanding and reasoning that are of great potential for both vision and Natural Language Processing (NLP). It allows linguistic constraints to be incorporated into semantic image segmentation (left), and  visual cues to improve entailment and paraphrasing (right).} \label{fig:overview}

\end{center}%
}]
\thispagestyle{empty}

\begin{abstract}
We introduce Segment-Phrase Table (SPT), a large collection of 
bijective associations 
between textual phrases and their corresponding segmentations. 
Leveraging recent progress in object recognition and natural language semantics, we show how we can successfully build a high-quality segment-phrase table using minimal human supervision. More importantly, we demonstrate the unique value unleashed by this rich bimodal resource, for both vision as well as natural language understanding. First, we show that fine-grained textual labels facilitate contextual reasoning that helps in satisfying semantic constraints across image segments. This feature enables us to achieve state-of-the-art segmentation results on benchmark datasets. Next, we show that the association of high-quality segmentations to textual phrases aids in richer semantic understanding and reasoning of these textual phrases. 
Leveraging this feature, we motivate the problem of visual entailment and visual paraphrasing,  and demonstrate its utility on a large dataset.
\end{abstract}


\vspace*{-4mm}
\section{Introduction}
\label{sec:intro}
\vspace*{-2mm}

Vision and language are among the primary modalities through which we humans acquire knowledge. A successful AI system would therefore need to leverage both these modalities together to arrive at a coherent and consistent understanding of the world. 
Although great progress has been made in well-constrained settings at the category level (e.g., objects, attributes, actions), several challenges (including scalability, modeling power, etc.,) have prevented us from bridging the gaps at higher levels. In recent years, impressive results have been demonstrated in aligning these two modalities at a much higher level~\cite{msrcaption,showtell,treetalk}, specifically, images to sentences, where an important enabling factor is the availability of large-scale image databases~\cite{ImageNet, Coco}, combined with deep learning methods~\cite{alexNet,jia2014caffe}.

The key ingredient in aligning vision and language involves reasoning about 
inter-modal correspondences~\cite{Farhadi10, Karapathy15,karpathy2014deep}.  
In this work, we target {\em phrase-level} associations, a relatively unexplored direction compared to category-level  and sentence-level associations. The goal is to achieve the specificity of category-level analysis while also maintaining the expressiveness of the high-level descriptions. 
A critical challenge is lack of large datasets with precise and exhaustive phrase-level alignments, as they are far more expensive to gather than sentence-level~\cite{Coco} or categorical labels~\cite{ImageNet}. 

Addressing this challenge, we present an approach, involving minimal human supervision, to build a high-quality \emph{Segment-Phrase Table (SPT)}, i.e., a large collection of bijective associations between phrases (e.g., `horse grazing') and their corresponding segmentations. 
Segments facilitate richer and more precise understanding of a visual phrase, e.g., the subtle difference between a `grazing horse' and a `standing horse' can be better captured via segmentation (fig.~\ref{fig:overview}). At the same time, textual phrases are commonly used semantic units in various state-of-the-art NLP systems (e.g. machine translation, paraphrasing). 
In this paper, we show that the large scale bi-modal associations between segments and phrases in SPT is of great value on important vision (e.g., semantic segmentation) and NLP (e.g., entailment, paraphrasing) tasks. 

\vskip .1cm
\noindent{\bf Segment-Phrase Table (SPT). } 
We present a scalable solution for obtaining an image-to-text translation dictionary between segments and  textual phrases.
We pursue an unsupervised approach,
analogous to how phrase-level translation tables are obtained in NLP without direct supervision on the internal semantic alignments. 

The core problem in building a segment-phrase table involves obtaining a segmentation mask (or model) for any given textual phrase. Conventionally, segmentation has been posed as a supervised learning problem where the training process requires images with detailed pixel-level annotations (indicating the extent of the foregrounds). 
Due to the high cost of pixel-level labeling, supervised segmentation is difficult to scale, especially for the rich space of textual phrases. 

We therefore introduce a method that leverages recent successes in webly-supervised object localization~\cite{Divvala14,sadeghi2015viske,chen2013neil} and co-segmentation~\cite{Joulin12,gunhee_cvpr12} for learning segmentation models directly from web images, using minimal human supervision. The core insight behind our approach is to cluster, localize, and segment instances by reasoning about the commonalities (in both semantic as well as appearance space) within a group of images retrieved for a given phrase. While our approach can be construed as akin to {\em Object Discovery}, we are faced with an additional challenge that the images being considered in our setting are a large collection of noisy (high intra-class variance) web images (i.e., without even image-level curation). To address this challenge, we employ a novel latent learning procedure that leverages cross-image consistencies within appearance-aligned instance clusters, thereby being robust to noise. To evaluate the quality of the segment-
phrase 
table we test it 
on  
object discovery datasets (Section.~\ref{sec:SPT}). 

As demonstrated in our experiments, the segment-phrase table facilitates bi-modal and rich semantic understanding and reasoning that are of great potential value for both vision and NLP tasks. More specifically, we show that the segment-phrase table allows (a) linguistic constraints to help image segmentation  (Section.~\ref{sec:seg}); (b) visual cues to improve entailment and paraphrasing  (Section.~\ref{sec:entail}). 

\vskip .1cm
\noindent {\bf Linguistic Constraints Help Semantic Segmentation.}  
Contextual knowledge has been shown to be valuable in improving the output label consistency in image segmentation~\cite{ChenJGD11}. For example, a `horse' segment is less likely to co-occur with a `microwave' segment. 
However, almost all previous approaches have limited their analysis to a small number of coarse (basic-level) categories and their small contextual relations. In this paper, we show how large scale rich segment-phrase associations in SPT allow us to take contextual reasoning to the next level (web-scale) and obtain state-of-the-art large-scale unsupervised image segmentation results by enforcing label consistencies (Section.~\ref{sec:seg}).   

\vskip .1cm
\noindent {\bf Visual Reasoning Helps Entailment and Paraphrasing.} 
The ability to read between the lines, and infer what is not directly stated has been a long standing challenge in many practical NLP tasks. Textual entailment, as studied in the NLP community, aims to address this challenge. More formally, given a pair of textual descriptions $X$ and $Y$,  $X$ entails $Y$ (i.e., $X \vDash Y$) if a human who reads $X$ would most likely infer $Y$ to be true. For example, `horse grazing' entails `horse standing'. Bidirectional textual entailment, known as paraphrasing in NLP, implies semantic equivalence. For example, bird flying is a paraphrase of bird gliding. Both these tasks are important constituents in a wide range of NLP applications, including question answering, summarization, and machine translation~\cite{surveyentail}. 

Previous approaches to textual entailment and paraphrasing have explored a variety of  syntactic and semantic cues, but most were confined to textual information. 
However, reasoning about the phrases in the visual domain offers crucial information not easily accessible in the text alone \cite{julia-entail}. For example, to determine whether the phrase `grazing horse' entails `standing horse', the visual similarity between them conveys valuable cues (figure.~\ref{fig:overview}). In fact, much of trivial common sense knowledge is rarely stated explicitly in text \cite{LoBue:2011:TCK:2002736.2002805}, and it is the segment-based visual analysis that can help filling in such missing information. 
Indeed, several recent approaches have explored how semantic representation of natural language can be enhanced when multimodal evidences are combined \cite{Q14-1017,angeliki15,julia-entail}. Our work contributes to this newly emerging line of research by presenting a segment-phrase table that draws mid-level correspondences between images and text, and shows its benefit for visual entailment and paraphrasing (Section.~\ref{sec:entail}). 

In summary, our key contributions are: (i) we motivate segment-phrase table as a convenient inter-modal correspondence representation and present a simple approach that involves minimal human supervision for building it. (ii) We demonstrate its use for the task of semantic segmentation, achieving state of the art results on benchmark datasets. (iii) We introduce the problem of visual entailment and paraphrasing, and show the utility of the segment-phrase table towards improving entailment and paraphrasing performance.

\section{Building the Segment-Phrase Table}
\label{sec:SPT}

\begin{figure}[t]
\centering
\includegraphics[width=\columnwidth]{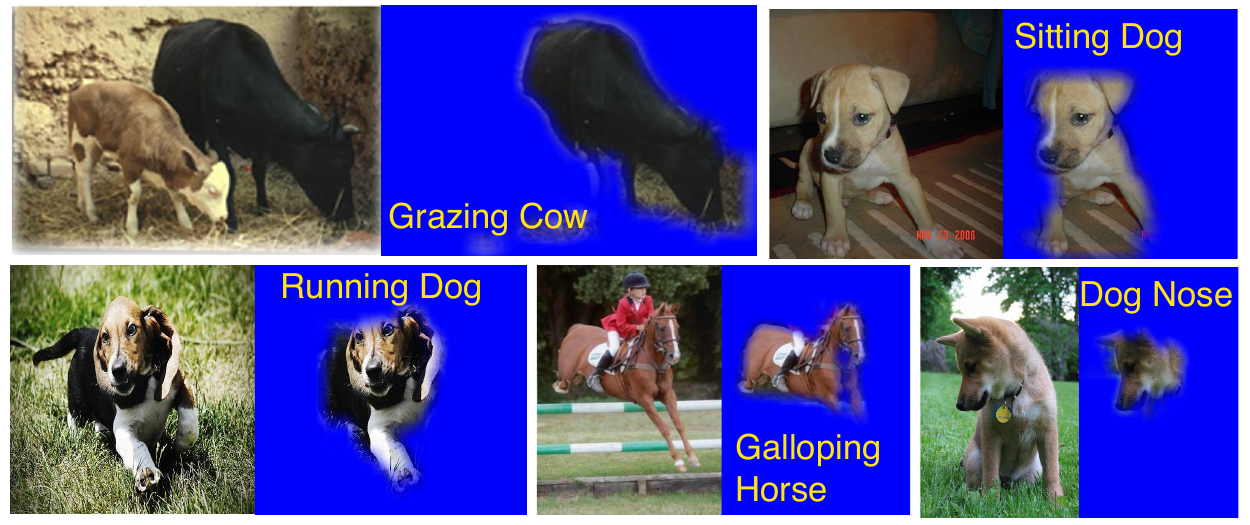} \vskip -.2cm
\caption{\small Examples of segments and their corresponding phrases from
the SPT. Notice the rich variety of phrases including parts (`dog
nose'), actions (`grazing cow', `galloping horse'), etc.}
\vspace{-4mm}
\label{fig:teaser}
\end{figure}

The key concern in building a Segment-Phrase Table (SPT) is scalability. Given that the expected size of the table would be potentially huge, it would be preferable to have an approach that requires no manual intervention and is computationally non-intensive, yet producing high-quality segmentation models. In the following, we present our method to generate the segment-phrase table. Figure~\ref{fig:teaser} shows examples of instances in SPT. 

Given an input textual phrase (e.g., `horse grazing'), the primary goal here is to obtain a segmentation model (mask) for that phrase. Given the recent popularity of segmentation-via-detection methods~\cite{sds_malik14}, we adopt a similar strategy in this work. Towards this end, we start with the webly-supervised approach of~\cite{Divvala14}, which focused on the task of detection, and extend it for the task of segmentation. 

In~\cite{Divvala14}, images retrieved from Internet are used to obtain well-performing object detection models for any given query phrase, while involving no manual intervention. Their method trains a mixture of $m$ components DPM~\cite{Felzenszwalb10} $M=(M_1,\hdots,M_m)$ in the weakly-supervised setting, where $M_c$ denotes the detection model for the $c$th component. Weakly-supervised localization is more manageable in their setting as the intra-class variance (within the downloaded images) per query is constrained. For example, images of `running horse' are more constrained than images of `horse' as the latter would have images of not only `running horse', but also `jumping horse', `grazing horse', `fighting horse', etc. This benefit leads to mixture components $M_c$ that are tightly clustered in the appearance space. For example, component $M_1$ of `running horse' ends up having all `running horse' frontal views, component $M_2$ would have `running horse' left views, etc.

Given a component $M_c$ and the training images $I_c$ along with their localized bounding boxes $B_c$, the next step is to learn a segmentation model $\theta^{fg}_c, \theta^{bg}_c$ for that component, where $\theta^{fg}$ indicates the foreground parameters and $\theta^{bg}$ indicates the background parameters. Since each component $M_c$ has low intra-class variance, we formulate our learning procedure as an iterative graph-based model that begins with the coarse bounding box-based correspondences across the instances, and refines them to yield fine segment-based correspondences. The key insight here is to leverage the cross-image consistencies within appearance-aligned instance clusters. 

Each image is modeled as a weighted graph $G= \{ V, E \}$, where $V =\{1,2,\hdots,n\}$  is the set of nodes denoting super-pixels~\cite{Meyer01}, and $E$ is the set of edges connecting pairs of adjacent super-pixels. For each node $i \in V$, a random variable $x_i$ is assigned a binary value $\{0,1\}$, building a Markov Random Field (MRF) representation. An energy function is defined over all possible labellings ${\bf x}=(x_1,\hdots,x_n) \in 2^n$~\cite{graphcuts} as 
\begin{equation} 
\label{gbeqn}
E({\bf x}) = \sum_{i \in V} u_i (x_i) + \sum_{(i,j) \in E} v_{ij}(x_i, x_j).
\end{equation}
The unary potential $u_i$ measures the agreement between the labels ${\bf x}$ and the image, while the pairwise potential $v_{ij}$ measures the extent to which ${\bf x}$ is piecewise smooth. The segmentation is obtained by minimizing the energy function using graph-cuts~\cite{graphcuts}. 

\vskip .1cm
\noindent {\bf Latent Learning.} To learn the model parameters, we need the pixel-level labeling of our training instances. However our training images $I_{c}$ are only accompanied with the bounding boxes $B_{c}$. To address the lack of supervision, the parameter learning task is formulated as a latent optimization problem where the pixel labels are treated as latent variables. We use an iterative expectation-maximization (EM) based method. Given the latent labeling of the pixels, the MRF parameters can be obtained using any supervised learning method. We use a simple Gaussian mixture model (GMM) trained on SIFT features (extracted within each superpixel) for modeling the unary parameters $\mu, \Sigma$, and set the unary potential to the GMM probability i.e., $u_i(x_i) = \frac{1}{\sqrt{(2\pi)^k|\Sigma|}} \exp(-\frac{1}{2} (f_i-\mu)^T \Sigma (f_i-\mu))$ ($f_i$ denotes the SIFT feature representation of $x_i$). The pairwise potential is defined based on the 
generalized boundary (Gb) detection probability~\cite{Leordeanu14} between two neighbouring superpixels $x_i, x_j$ i.e., $v_{ij} = \exp(-\lambda P_{Gb}(x_i,x_j))$ ($\lambda=0.05$). In the E-step, given the model parameters, the labels for the pixels are inferred by running Graph-cut~\cite{graphcuts}. 

\vskip .1cm
\noindent {\bf Initialization.} To initialize our latent learning method, we apply Grabcut~\cite{grabcut} on each training instance. Grabcut helps in obtaining a rough, noisy foreground-background labeling of pixels. We found this to be a reasonable initialization in our experiments. Grabcut in turn needs weak/interactive annotation for initialization. We use the HOG template model associated with the component to guide the Grabcut initialization. Given the HOG template model $M_c$, we compute the regions of high energy within the root filter, where energy is defined as the norm of the positive weights. All superpixels within this high energy regions are initialized to foreground and the rest as background. We found this initialization to produce good results in our experiments.

\section{Semantic Segmentation using Linguistic Constraints}
\label{sec:seg}

There is a  consensus that semantic context plays an important role of reducing ambiguity in objects visual appearance. When segmenting an object or a scene, the category labels must be assigned with respect to all the other categories in the image. Based on this insight, several methods have been proposed to incorporate contextual relations between objects’ labels~\cite{objectsincontext,divvala2009empirical,izadinia2014incorporating,ChenJGD11}. However, to our knowledge, there does not exist an unsupervised  approach in literature that incorporates large-scale semantic contextual constraints explicitly at the phrase-level for enforcing output label consistency. The lack of annotations coupled with the noise in the Web images presents a herculean challenge to reason about large-scale contextual relationships in the unsupervised setting. In this work, we show that it is possible to incorporate such contextual constraints by leveraging our segment-phrase table.

Recall that each segmentation model $\theta^{fg}, \theta^{bg}$ in our segment-phrase table is associated with a textual phrase. For any given object category (e.g., `horse'), we consider all the segmentation models belonging to its related phrases $p \in V=\{v_1,\hdots,v_N\}$.\footnote{The phrases related to a category are obtained using the query expansion procedure of~\cite{Divvala14}.} 

To segment a test image, we first run the detection models $M$ per phrase $p$ (denoted as $M^p$) and localize all its possible variations. In order to ensure high recall, we perform non-maxima suppression on the detections of each phrase independently, and pick all top detections above a threshold. For each detection window, we use its corresponding segmentation model from SPT to segment it. This segmentation step involves running the Graphcut inference step~\cite{graphcuts} using the  model parameters ($\theta^{fg}, \theta^{bg}$) in SPT. The output is a binary foreground-background mask for that detection window. We multiply this foreground mask by its corresponding detection score $S(M^p_c)$ (produced by the detector $M^p$) to obtain a weight for that mask. This process is repeated for each detection. For a given test image, this step results in a pool of weighted foreground masks each of which associated with a textual phrase. 

Given this pool of foreground masks, for example, `horse fighting', `horse rearing', `horse jumping', `horse lying', and `horse cap', etc., the labels corresponding to `fighting', `rearing' and `jumping' are in context (as fighting horses typically jump and rear) and therefore should reinforce each other, while `horse lying' and `horse cap' are out of context (probably due to some ambiguities in their visual appearance) and therefore should be suppressed. To determine such congruence of labels, we use distributed vector representations that capture syntactic and semantic word relationships. 

\vskip .1cm
\noindent {\bf Enforcing Linguistic Constraints.} Distributed representations of words in a vector space is popular in the NLP community to help learning algorithms achieve better performance at language analysis tasks that involve analyzing the similarity between words~\cite{word2vec}. The word representations are valuable as they encode linguistic regularities and patterns that can be represented as linear translations. 

In our work, we use these vector representations to compute similarity between two fine-grained semantic composite  labels. Given a composite label, such as `horse jumping', we use the approach of~\cite{Mikolov13} to obtain its vector representation: $V(\text{`horse jumping'}) = V(\text{`horse'}) \oplus V(\text{`jumping'})$. The $\oplus$ operator yields a vector originating at `horse' and pointing towards `jumping'. This aims at approximating what `jumping' does to `horse'. To estimate the similarity between any two composite labels $p,q$, we compute the cosine distance between their corresponding composite vector representations i.e., $\psi(p, q) = cosine(V(p), V(q))$. In our implementation, we use word2vec embeddings and represent labels using a 300 dimensional representation computed using the pre-trained vectors of~\cite{word2vec}.

To enforce label consistency using this linguistic prior, we construct a fully-connected graph using the pool of foreground masks, where the nodes are individual foreground segments, and the edges are phrase similarities $\psi$. We perform a single round of message passing over this graph to obtain the new scores for each of the masks. The new scores are computed as: $S(M^p_c) = \sum_q S(M^q_c) \times \psi(p,q)$.

Given these re-scored foreground masks, we use sum-pooling to merge the multiple masks and estimate a single resultant weighted mask for the image. We apply Graphcut using this weighted mask as the unary potential and  the same pairwise potential as used in Section~\ref{sec:SPT}. The output of this Graphcut is the final semantic pixel labeling of the test image (See Figure.~\ref{fig:imgnetseg} for examples of unsupervised segmentation of ImageNet). 

\section{Visual Semantic Relations}
\label{sec:entail}
The availability of rich bijective associations in our segment-phrase table representation opens the door for several new capabilities in natural language understanding and reasoning. In this work, we have analyzed its utility for three challenging NLP problems: entailment, paraphrasing, and relative semantic similarity. 

\subsection{Visual Entailment} 
\label{sec:ventail}
If we come across a fact that a horse is eating, then based on our common knowledge, we infer that the horse must be standing. Reasoning about such knowledge is the primay focus of the entailment task. More specifically, $X$ \emph{entails} $Y$, i.e., if $X$ is true, then $Y$ is most likely  (based on commonsense) also true. Positive entailments are denoted as $\vDash$, while non-entailments are denoted as $\nvDash$ (for e.g., \emph{`horse lying'} $\nvDash$ \emph{`horse standing'}). Entailment continues to be an open-research challenge in the NLP community, where its scope has been largely limited to textual domain~\cite{surveyentail}. In the following, we show that our SPT provides unique capabilities for inferring entailment in short (visual) phrases.

Let $\mathcal{T}(X)$ and $\mathcal{T}(Y)$ be the set of worlds in which $X$ and $Y$ are true respectively. From the set-theoretic perspective, $X \vDash Y$ iff. $\mathcal{T}(X) \subset \mathcal{T}(Y)$. We approximate $\mathcal{T}(X)$ and $\mathcal{T}(Y)$ by the set of images in which $X$ and $Y$ appear.  
If $X \vDash Y$, this means that for all segments for which $X$ can be used as a textual description, it must also be that $Y$ can be used as a description. However $X$ does not need to be applicable for all segments within $Y$. Intuitively, this means $Y$ can be applied in a more visually divergent context than $X$. To capture this intuition of directional visual compatibility, we define the visual entailment score as:
\begin{equation}\label{eq:segbasedentail}
 \mbox{entail}(X \vDash Y) := Sim_{R2I}^{\rightarrow}(X,Y) - Sim_{R2I}^{\rightarrow}(Y,X),
\end{equation}
where $Sim_{R2I}^{\rightarrow}$ refers to (directed) visual similarity measure between two phrases. To compute $Sim_{R2I}^{\rightarrow}(X,Y)$, we use the top K (K=10) segmentation masks corresponding to the phrases $X,Y$ in our SPT, and estimate the (average) similarity between these masks using the asymmetric region-to-image similarity measure introduced in~\cite{kim2010asymmetric}.

\vskip .1cm

\noindent {\bf Enforcing transitivity constraints via global reasoning.}
While it is possible to reason about an entailment relation $X \vDash Y$ independently using the above approach, the availability of large pool of (related) phrases in our SPT facilitates richer global reasoning that can help satisfy higher-order transitivity constraints. For example, the knowledge of \emph{`horse rearing'} $\vDash$ \emph{`horse jumping'}, and \emph{`horse fighting'} $\vDash$ \emph{`horse rearing'}, should influence when determining \emph{`horse fighting'} $\vDash$ \emph{`horse jumping'}. To enforce such transitivity constraints, we perform a global inference over an entailment graph $\mathcal{G}$. Each node of the graph represents an input phrase and an edge $e\in\mathcal{E}$ between two nodes represents their entailment relationship. For every two nodes $x, y\in \mathcal{V}$, we compute $\mbox{entail}_{xy}$  
(as in eq (\ref{eq:segbasedentail})) 
as the edge weight between them. Enforcing transitivity in the graph implies that if there exists edges $e(x,y)$ and $e(y,z)$ in the graph, then the edge $e(x,z)$ should also exist. In other words, whenever the relations $x \vDash y$, and $y \vDash z$ hold, we have $x \vDash z$.

We formulate this problem as an integer linear programming, where a linear function is maximized under linear constraints, with the goal of finding existence (or non-existence) of  entailment relations. To find the best entailment graph that has the maximum sum over the edge scores~\cite{Berant}, we need to find the weights $W_{xy}$ that determine whether edge $e(x,y)$ exists in the final entailment graph or not i.e., 

\begin{eqnarray}\label{ilp}
{\max} \hspace*{-5mm}&&\sum_{x\neq y}{\mbox{entail}_{xy}W_{xy} -\lambda|W|}  \ \ \  s.t.  \ \ \  W_{xy} \in \{0,1\},   \nonumber \\
 && \forall x,y,z \in \mathcal{V}, W_{xy} + W_{yz} - W_{xz} \leq 1
\end{eqnarray}
\noindent where $W_{xy} + W_{yz} - W_{xz} \leq 1$ ensures that the transitivity constraint is satisfied. 

\subsection{Visual Paraphrasing} 
\label{sec:vp}
Paraphrasing is a related NLP task where the goal is to recognize phrases with semantic equivalence. For example, the knowledge of a `horse jumping' is (semantically) same as that of a `horse leaping'. The paraphrasing task is often considered as a special case of entailment, where bidirectional entailment relationship holds. Based on this observation, we extend our proposed visual entailment approach to also address the problem of visual paraphrasing. 

In defining the notion of semantic equivalence, textual paraphrasing approaches adopt a relaxed notion as it is of greater practical value to detect paraphrases that are lexically divergent rather than those with trivial differences in wordings. Following this relaxed notion, we define a pair of phrases to be paraphrases if the absolute difference of the entailment scores ($\mbox{entail}_{xy}$ and $\mbox{entail}_{yx}$) is within a threshold. 

By leveraging the visual knowledge available in our SPT, we show that it is possible to find interesting paraphrases that have not been previously detected in the existing NLP resources (see Figure~\ref{fig:nlpqual} in Section~\ref{sec:nlpresults}).  

\subsection{Relative Semantic Similarity} 
\label{sec:rss}
Quantifying semantic similarity is one of the basic tasks in NLP. Although there exist many approaches and resources for measuring word-level analysis, relatively little exists for phrase-level.
However to determine whether `cat standing up' (in Fig.~\ref{fig:overview}) is visually more similar to `bear standing up' than to `deer standing up', word-level lexical similarities alone would not be sufficient. 
Our SPT provides a natural means to compute such semantic distances visually directly at the phrase-level. In our experiments, we use the entailment score measure $\mbox{entail}_{xy}$ as the semantic similarity between the phrases $x,y$.


\section{Experiments and Results}

\begin{table*}[t]
\centering
\begin{small}
\begin{tabular}{|l|c|cc|cc|cc||cc|cc|cc|}
\hline
 &  & \multicolumn{6}{c||}{Full set}  & \multicolumn{6}{c|}{Subset (100 samples per category) }\\
 \hline  & Train/Test & \multicolumn{2}{c|}{Car}  & \multicolumn{2}{c|}{Horse} & \multicolumn{2}{c||}{Plane} & \multicolumn{2}{c|}{Car}  & \multicolumn{2}{c|}{Horse} & \multicolumn{2}{c|}{Plane}\\
Method & same data & (P) & (J) & (P) & (J) & (P) & (J) & (P) & (J) & (P) & (J) & (P) & (J)\\
\hline
\cite{Joulin10} & yes & -- & -- & -- & -- & -- & -- & 58.7 & 37.15 & 63.84 & 30.16 &49.25 & 15.36\\
\cite{Joulin12} & yes & -- & -- & -- & -- & -- & -- & 59.2 & 35.15 & 64.22 & 29.53 & 47.48 & 11.72\\
\cite{gunhee_cvpr12}& yes   & -- & -- & -- & -- & -- & -- & 68.85 & 0.04 & 75.12 & 6.43 & 80.2 & 7.9\\
\cite{celiu13}& yes  & 83.4, & 63.4 & 83.7 & 53.9 & 86.1 & 55.6 & 85.38 & 64.42 & 82.81 & 51.65 & 88.04 & 55.81\\
\cite{Chen14} & yes & 87.1 & 64.7 & 89 & 57.6 & 90.2 & {\bf 60.0} & 87.65 & 64.86 & 86.16 & 33.39 & {\bf 90.25} & 40.33 \\
\cite{Divvala14} baseline & no & 85.7 & 69.1 & 78.7 & 44.2 & 72.2 & 31.3 & 85.41 & 70.04 & 77.39 & 41.58 & 71.59 & 31.22 \\
SPT (ours) & no & {\bf 88.2} & {\bf 69.9} & {\bf 91.2} & {\bf 59.0} & {\bf 91.0} & 57.1 & {\bf 88.27} & {\bf 70.14} & {\bf 89.01} & {\bf 54.12} & 88.85 & {\bf 56.19} \\
\hline
\end{tabular}\vskip -.2cm
\end{small}
\caption{\small Segmentation results on Bing dataset~\cite{celiu13}. Our approach outperforms the previous methods on both the precision and Jaccard measures. Row 6 demonstrates the result obtained by turning off the latent learning procedure in our approach i.e., learning the model parameters directly from the grabcut initialization over the detections of~\cite{Divvala14}. Latent learning indeed contributes towards considerable improvement in performance.}
\label{tab:bing}
\end{table*}

\begin{table*}[t]
\centering
\begin{tabular}{|c|l|c|c|c|c|c|c|c|c|c|c|c|}
\hline
 & Method  & Supervision & aero  & boat  & car   & cat   & cow   & dog   & horse & sheep & tv    & mean  \\
\hline
 & baseline \cite{Divvala14}   & no      &  73.8 & 77.5 & 72.9 & 72 & 78.2 & 75.6 & 76.2 & 77.5 & 76.3 & 75.5 \\
 Precision & SegProg~\cite{Kuettel12} & \textbf{yes} & 87.2 & 85.3 & 82.8 & 82.3 & 85.5 & 83.9 & 81.7 & 81.8 & 80.2 & 83.4 \\
 & SPT (ours)                & no &  75.4 &  78.5 & 84.2 & 78.3 & 83.8 & 83.3 & 78.5 & 82.6 & 80.2 & 80.5 \\
\hline
\hline
 & baseline \cite{Divvala14}    & no   &  32.7 & 31.9 & 48.6 & 33 & 45.5 & 38.4 & 22.2 & 43.7 & 48.9 & 38.3 \\
 Jaccard & SegProg~\cite{Kuettel12} & \textbf{yes} & 59.5 & 48.6 & 66.9 & 61.9 & 61.8 & 61.3 & 49.1 & 55.9 & 61 & 58.5 \\
 & SPT (ours)             & no   &  43.5 & 44.7 & 71.5 & 50.1 & 58.7 & 58.8 & 46.4 & 55.9 & 59.9 & 54.4 \\
\hline
\end{tabular}\vskip -.2cm
\caption{\small Segmentation Results on ImageNet Auto-annotation: We compare our method with the method of~\cite{Kuettel12} and a baseline based on \cite{Divvala14}. while~\cite{Kuettel12} uses Pascal segmentation masks, ImageNet bounding boxes, and WordNet taxonomy, our model needs no explicit supervision. Nonetheless, it obtains comparable results.}
\label{tab:imgnet}
\end{table*}

We evaluate the quality of the segment-phrase table  and its application in semantic segmentation, where language helps vision, and visual entailment and paraphrasing, where vision helps language.

\subsection{SPT Quality Evaluation}
\label{sec:spt_res}

We evaluate the quality of our segment-phrase table on standard segmentation datasets. Our results show that the SPT   (i.e., even without enforcing the linguistic constraints) performs on par with other state-of-the-art methods.

\vskip .1cm
\noindent {\bf Joint Object Discovery and Segmentation.} In~\cite{celiu13}, the problem of automatically discovering and segmenting out the common regions across a pool of images for an object category was studied. We analysed the performance of the segment-phrase table on this task.

Table~\ref{tab:bing} compares the results of our approach with other competing baselines on this dataset~\cite{celiu13}. Quantitative evaluation is performed against manual foreground-background segmentations that are considered as `ground truth'. We use similar  performance metrics: precision P (ratio of correctly labeled foreground and background  pixels), and Jaccard similarity J (intersection divided by the union of the segmentation result with the ground truth segmentation).

\begin{figure}[t]
\centering
\includegraphics[scale=0.6]{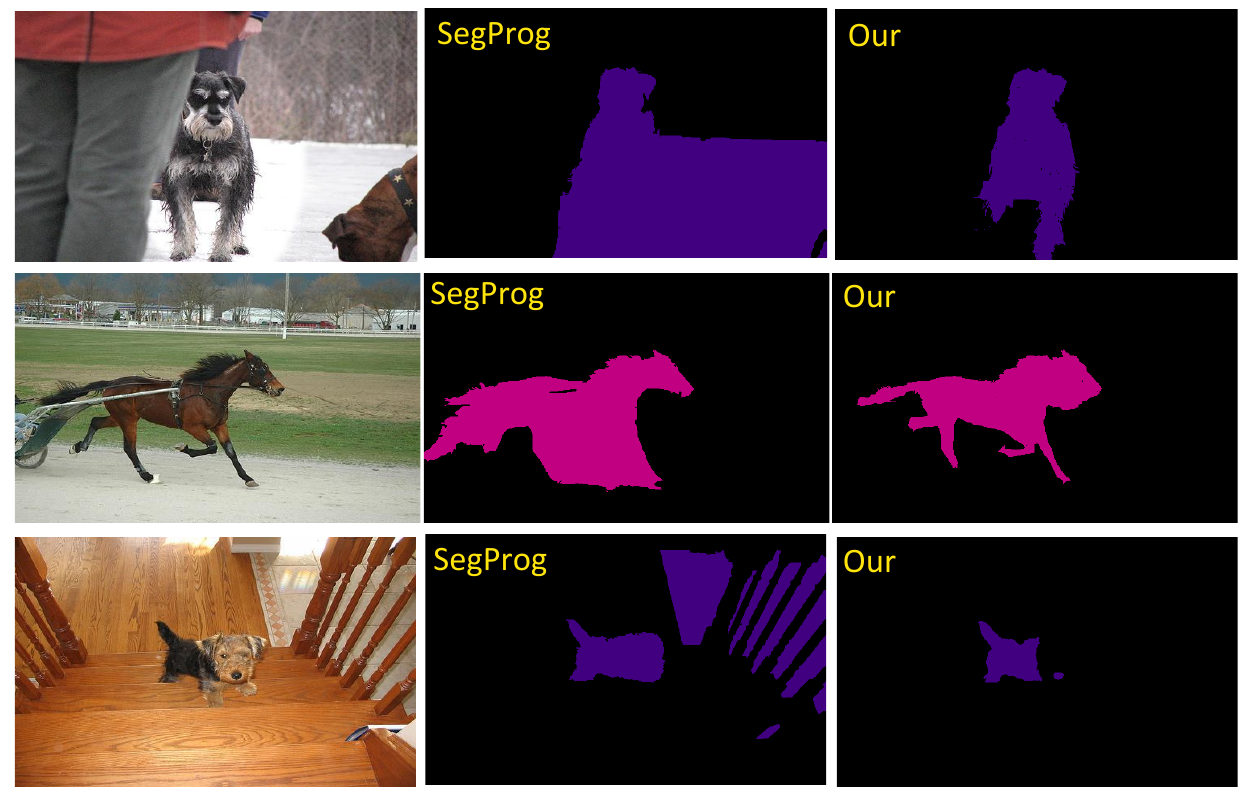}\vskip -.2cm
\caption{\small Results obtained on the ImageNet auto-annotation task. The approach of~\cite{Kuettel12} uses Pascal segmentation masks, ImageNet bounding boxes, and WordNet taxonomy, while our approach uses no explicit supervision.}
\vspace{-5mm}
\label{fig:imgnetseg}
\end{figure}

Our approach outperforms previous methods on both the metrics. This strong result brings to light an interesting trade-off: using large datasets with zero human supervision vs. smaller datasets with some human supervision. Previous works have studied the latter setting, where the number of images considered were small but had some curation. For example, \cite{Joulin10,Joulin12} uses MSRC \& iCoseg, \cite{gunhee_cvpr12} uses FlickrMFC. Using the segment-phrase table, our approach exposes the former setting i.e. using a larger pool of web images but with zero curation. The amount of human supervision used in making these datasets (MSRC, iCoseg, etc.) is an important aspect when scalability is of concern.

In comparison to the method of~\cite{celiu13}, the segment-phrase table not only obtains better results, but is also computationally attractive. \cite{celiu13} requires a measure of visual saliency and is thereby slow due to the estimation of dense correspondences (using SIFT flow). As a result, it considers a medium-scale imageset (4000 images) per category. However, in our approach as the segment-phrase table representation facilitates meaningful organization of the data (both in the semantic as well as appearance space), the intra-class variance is highly constrained. This feature simplifies the cosegmentation task and therefore allows the use of simple models to achieve good results.

\vskip .1cm
\noindent {\bf ImageNet Auto-annotation.} In~\cite{Kuettel12}, the problem of automatically populating the ImageNet dataset with pixelwise segmentations was studied. A unique feature of our segment-phrase table representation is that generic models can be learned (in an offline setting) and be used to perform segmentation across multiple datasets. This feature circumvents the limitation of previous methods~\cite{celiu13,Chen14} that operate in the transductive setting (where the test and training sets are identical). Given this benefit, we analyze the performance of our models on the auto-annotation task as well.

We follow the experimental set up of~\cite{Kuettel12}\footnote{The ground-truth masks and the foreground masks produced by the segmentation propagation method of~\cite{Kuettel12} are both available on their project website, enabling us to compare our results.} and use a similar procedure as detailed earlier for segmenting the images. Table~\ref{tab:imgnet} compares our results with~\cite{Kuettel12}. Our method obtains similar levels of precision with a small drop in Jaccard. Note that our method does not require any form of manual supervision, while~\cite{Kuettel12} uses a (i) seed set of segmentation masks (VOC2012), (ii) a taxonomy structure (WordNet), and (iii) a held-out set of labeled images (with bounding boxes) to train their model parameters. Figure~\ref{fig:imgnetseg} shows some qualitative examples of comparisons between our method and \cite{Kuettel12}. 

\noindent {\bf Qualitative Results.} Using our approach, we have produced a segment-phrase table containing over 50,000 entries. Figure~\ref{fig:teaser} shows some examples, covering a variety of phrases including parts (`dog nose'), attributes (`briddle horse'), actions (`grazing cow'), etc. We believe this table has the largest number of phrasal segmentation models reported in the literature. 

\subsection{SPT for Semantic Segmentation}

In Table~\ref{tab:bing_slc}, we analyze the effect of enforcing linguistic constraints on the above tasks. Introducing linguistic constraints results in a significant gain in performance. This result shows the value of using semantic constraints to improve segmentation accuracy (by way of enforcing output label consistency). Our work demonstrates the possibility of enforcing such semantic constraints in an unsupervised manner by leveraging the availability of phrasal labels in our segment-phrase table. 
\begin{table}[t]
\centering
\begin{small}
\begin{tabular}{|l|cc||cc||cc|}
\hline
  & \multicolumn{2}{c||}{Bing (Full)}  & \multicolumn{2}{c||}{Bing (Subset)} & \multicolumn{2}{c|}{ImageNet} \\
Method & (P) & (J) & (P) & (J) & (P) & (J)\\
\hline
SPT & {\bf 90.1} & 62.0 & 88.7 & 60.2 & 80.5 & 54.4 \\
SPT+L & 89.7 & {\bf 65.9} & {\bf 89.3} & {\bf 63.3} & {\bf 82.5} & {\bf 55.7} \\
\hline
\end{tabular}\vskip -.2cm
\end{small}
\caption{\small Enforcing linguistic constraints helps semantic segmentation. This table displays the results with   (bottom row) and without (top row) enforcing linguistic constraints. The mean performance (Precision and  Jaccard) across all the categories on the joint object discovery and segmentation task in Bing (Full) and Bing (Subset) as well as the ImageNet auto-annotation task is reported. }
\label{tab:bing_slc}
\end{table}
\subsection{SPT for Visual Semantic Relations}\label{sec:nlpresults}

\begin{figure*}
\centering
\includegraphics[scale=0.44]{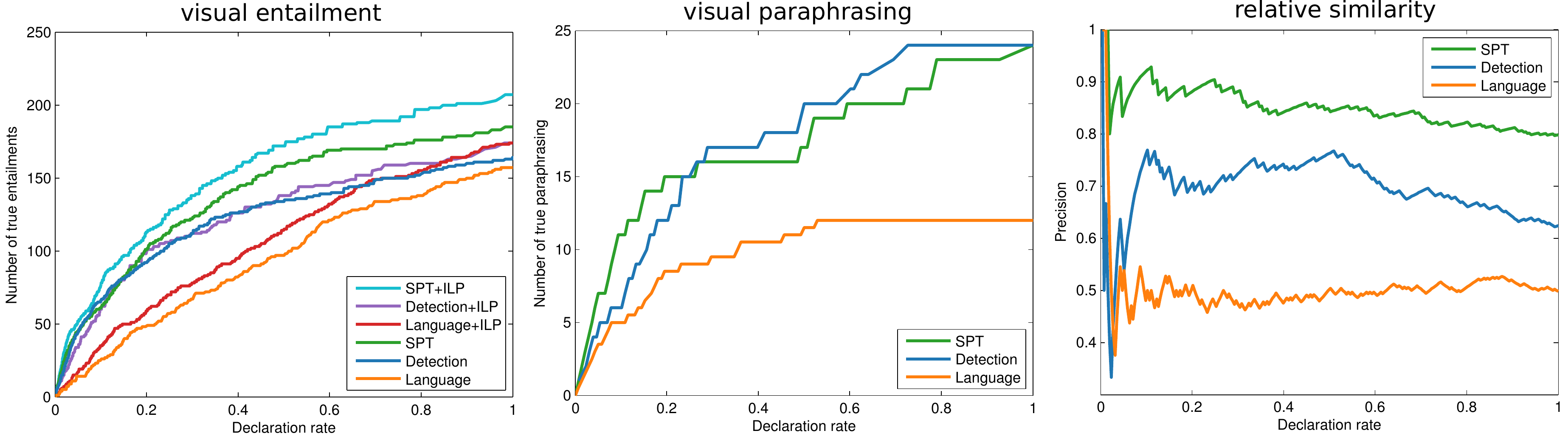}\vskip -.3cm
\caption{\small Visual Semantic Relation Results. Our SPT-based approach
obtains better results compared to baseline methods. In case of the entailment task, global reasoning further
helps in improving the performance.}\vskip -.2cm
\label{fig:nlpres}
\end{figure*}

\begin{figure}
\centering
\includegraphics[scale=0.29]{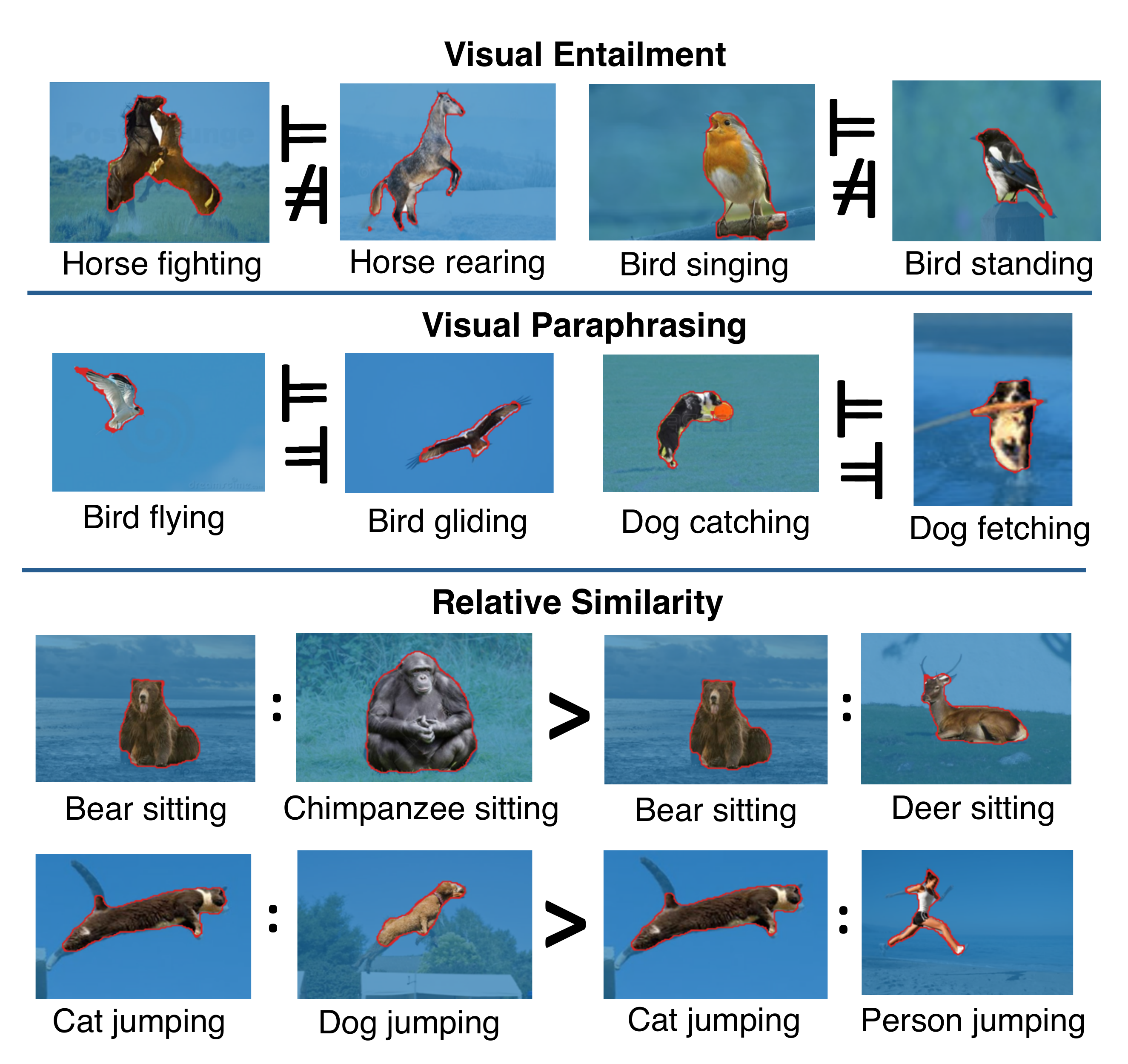}\vskip -.3cm
\caption{\small Qualitative examples for visual entailments, paraphrasing
and relative similarities inferred by our method. }
\vspace{-3mm}
\label{fig:nlpqual}
\end{figure}
We evaluate applications of SPT to three main language tasks: entailment, paraphrasing, and also relative similarities. For each of these tasks we have a detection-based baseline to show the importance of segments in the SPT, and a language-based baseline to show the gain in using rich visual information encoded by SPT. For qualitative results, please refer to Figure~\ref{fig:nlpqual}. 

\vskip .1cm
\noindent{\bf Dataset and Metrics:} Building upon the relation phrases of~\cite{sadeghi2015viske}, we collected a dataset of 5,710 semantic relations that include entailment (270 true relations), and paraphrasing (25 true relations). We also collected a dataset of 354 relative similarity statements that contains 253 valid relative similarity statements. We report number of correctly labeled semantic relations versus the declaration rate as our evaluation metric. 
\subsubsection{Visual Entailment}\label{sec:expentail}
\noindent{\bf{Detection-based Entailment Baseline.}}
To provide an apple-to-apple comparison between webly supervised segmentation and bounding boxes for visual text entailment, we also define entailment score based on detected bounding boxes. We use almost the same scheme as shown in Section~\ref{sec:entail} to compute $\mbox{entail}_{xy}$  except that here the bounding boxes obtained from DPM detections (instead of segments from SPT) are used to compute the similarity between two images. Figure~\ref{fig:nlpres} reports results for this baseline (`Detection') as well as a version that this baseline is augmented by the ILP formulation~\ref{ilp} (`Detection+ILP').

\vskip .1cm
\noindent{\bf{Language-based Entailment Baseline.}} 
There is little prior work for inferring textual entailment for short phrases. 
We therefore propose a new method for text-based entailment that is analogous to the vision-based entailment method described above. We obtain  the textual semantic representation of  phrases $X$ and $Y$  by element-wise addition of  word2vec embeddings\cite{word2vec} of the words in each phrase.\footnote{Element-wise multiplication is another common choice. In our experiments particular choice of vector composition did not yield much difference. We only report results based on element-wise addition for brevity.} We compute contextual similarity as the cosine distance between the phrase embeddings. 

We assign $X \vDash Y$ to be true only when their contextual similarities are bigger than a threshold. Although contextual similarity differs from the notion of entailment, many previous studies found that contextual similarities to be one of the useful features for recognizing entailment~\cite{surveyentail}. Since most pairs are not in the entailment relations, it is possible to build a competitive baseline by predicting entailment only when there is a strong contextual support. 
To determine the directionality of entailment, we  compare the size of the corresponding possible worlds. We approximate this by comparing probability scores obtained based on MSR Language Model \cite{Huang10}.  

We also considered the use of word-level entailment resources such as WordNet~\cite{wordnet} and VerbOcean~\cite{verbocean}, but we found that both resources contain very sparse entailment relations to produce any competitive baseline for our visual phrase entailment task. In addition, word-level entailment does not readily propagates to phrase-level entailment. For example, while `horse eating' would entail `horse standing', `cat eating' does not entail `cat standing'. We report results both for this baseline (`Language') and augmented by the ILP formulation (`Language+ILP') in Figure~\ref{fig:nlpres}.

\vskip .1cm
\noindent{\bf Visual Entailment Results.}
Figure~\ref{fig:nlpres} shows comparison of our method with the baselines.  Visual information encoded by SPT greatly improves the entailment results compared to a competitive language baseline.  It is also interesting to note that the SPT-based entailment outperforms the detection-based baseline confirming our intuition that segments are necessary for subtle visual reasoning. The best performance is achieved when the global inference is applied over the entailment graph for enforcing transitive closure.

\vskip -.1cm
\subsubsection{Visual Paraphrasing}
\noindent{\bf Baselines.} We use the language-based and detection baselines as explained in Section~\ref{sec:expentail}. The language-based baseline only computes contextual similarities (cosine similarity between phrase embeddings) and does not use the language model frequencies. The detection baseline is similar to our method where the paraphrasing  score is computed based on the bounding boxes rather than segments.

\vskip .1cm
\noindent{\bf Visual Paraphrasing Results.} Figure~\ref{fig:nlpres} shows comparison of our method in visual paraphrasing with the baselines.   Our SPT-based paraphrasing significantly outperforms  the language-based baseline.

\subsubsection{Relative Visual Similarity}
\noindent{\bf Baselines.} We use language-based and detection baselines similar to the two previous task. For every baseline, we compute the similarity between two phrases and report the one with higher similarity.

\vskip .1cm
\noindent{\bf Relative Visual Similarity Results.} Figure~\ref{fig:nlpres} shows comparison of our method in  relative visual similarity with the baselines.  Our SPT-based approach outperforms the competitive baselines. It is interesting to see in Figure~\ref{fig:nlpqual} that SPT enables rich and subtle visual reasoning that results in inferences such the way `bears' sit is more similar to that of `chimpanzees' compared to `deers' sitting.

\section{Conclusion}
\label{sec:concl}

In this work, we have introduced the {\em segment-phrase table} as a large collection of bijective associations between textual phrases and their corresponding segmentations. Such a representation achieves the specificity of category-level analysis while maintaining the expressiveness of the high-level descriptions. We have demonstrated the great value offered by this resource for both vision as well as natural language understanding tasks. By achieving impressive results on the segmentation task, we have shown that the segment-phrase table allows leveraging linguistic constraints for semantic reasoning. The association of high-quality segmentations to textual phrases also aids richer textual understanding and reasoning. We have shown its benefit on the tasks of entailment, paraphrasing and relative similarity. Our approach enables interesting research directions and opens the door for several applications.

\paragraph{Acknowledgments:} This work was supported by ONR N00014-13-1-0720, NSF IIS-1218683, NSF IIS-IIS-1338054, and Allen Distinguished Investigator Award.

{\small
\bibliographystyle{ieee}
\bibliography{global}
}

\end{document}